\RequirePackage{fix-cm}
\documentclass{svjour3}                  
\smartqed  
\usepackage{CJKutf8}
\usepackage{lineno}
\usepackage{footnote}
\usepackage{hyperref}
\usepackage{tablefootnote}
\usepackage{tabularx}
\usepackage[T1]{fontenc}
\usepackage{graphicx}
\usepackage{adjustbox}
\usepackage{gensymb}

\usepackage{epsfig}
\usepackage{wrapfig}
\usepackage[ruled]{algorithm2e}

\usepackage{multirow}

\begin{document}

\title{Stylometry Analysis of Human and Machine Text for Academic Integrity}

\author{\mbox{Hezam Albaqami}   \and
        \mbox{Muhammad Asif Ayub}         \and
        \mbox{Nasir Ahmad}         \and  
	    \mbox{Yaseen Ahmad}         \and
         \mbox{Mohammed M. Alqahtani}  \and  
        \mbox{Abdullah M. Algamdi}  \and
        \mbox{Almoaid A. Owaidah}  \and
        \mbox{Kashif Ahmad}
       }
\institute{Hezam Albaqami \at
             Department of Computer Science and Artificial Intelligence, College of Computer Science and Engineering, University of Jeddah, Jeddah 21493, Saudi Arabia. \\
              \email{haalbaqamii@uj.edu.sa}           
           \and
           Muhammad Asif Ayub \at
              University of Engineering and Technology, Peshawar, Pakistan. \\
              \email{asifayub836@gmail.com}
             \and
                Nasir Ahmad \at 
              Munster Technological University, Cork, Ireland. \\ \email{nasir.ahmad@mtu.ie}
              \and
              Yaseen Ahmad \at 
              University of Engineering and Technology, Peshawar, Pakistan. \\ \email{ahmadyaseen1628@gmail.com}
              \and
             Mohammed M. Alqahtani \at 
              Department of Computer Science and Artificial Intelligence, College of Computer Science and Engineering, University of Jeddah, Jeddah 21493, Saudi Arabia. \\ \email{Mmalqahtani@uj.edu.sa}
              \and
               Abdullah M. Algamdi \at 
              Department of Computer Science and Artificial Intelligence, College of Computer Science and Engineering, University of Jeddah, Jeddah 21493, Saudi Arabia. \\ \email{amalgamdi6@uj.edu.sa}
              \and
               Almoaid A. Owaidah \at 
              Department of Management Information Systems, Faculty of Economics and Administration, King Abdulaziz University, Jeddah, Jeddah 21589, Saudi Arabia. \\ \email{aowaidah@kau.edu.sa}
              \and
                Kashif Ahmad \at 
               Munster Technological University, Cork, Ireland \\ \email{kashif.ahmad@mtu.ie}
              \and
              }

\maketitle
\begin{abstract}
This work addresses critical challenges to academic integrity, including plagiarism, fabrication, and verification of authorship of educational content, by proposing a Natural Language Processing (NLP)-based framework for authenticating students' content through author attribution and style change detection. Despite some initial efforts, several aspects of the topic are yet to be explored. In contrast to existing solutions, the paper provides a comprehensive analysis of the topic by targeting four relevant tasks, including (i) classification of human and machine text, (ii) differentiating in single and multi-authored documents, (iii) author change detection within multi-authored documents, and (iv) author recognition in collaboratively produced documents. The solutions proposed for the tasks are evaluated on two datasets generated with Gemini using two different prompts, including a normal and a strict set of instructions. During experiments, some reduction in the performance of the proposed solutions is observed on the dataset generated through the strict prompt, demonstrating the complexities involved in detecting machine-generated text with cleverly crafted prompts. The generated datasets, code, and other relevant materials are made publicly available on GitHub\footnote{https://github.com/MASIFAYUB/Stylometry-Analysis-of-Human-and-Machine\#}, which are expected to provide a baseline for future research in the domain.

\end{abstract}

\keywords{Author Attribution \and Author Detection \and Style Change Detection \and Machine Text Detection \and Generative AI \and Text Classification}

\section{Introduction}
In recent years, an enormous interest has been observed in Generative Artificial Intelligence (AI) and its applications. As a result, several interesting AI-based content/text-generating tools, such as ChatGPT, have been introduced that can accurately generate grammatically correct text on demand for various purposes \cite{fariello2025distinguishing}. For instance, articles, emails, social media posts, essays, and even code in different programming languages could be generated on the fly in response to questions asked by the users. These auto-text generating tools are useful in multiple domains, particularly in the education sector, where they provide diverse knowledge sources with reduced bias compared to conventional learning approaches. However, if used unethically, these tools could significantly harm the learning outcomes and objectives of education. For instance, students may rely on them to complete assignments without fully engaging with or understanding the underlying questions. The plagiarism and unethical use of these tools in assignments, tests, thesis, and project reports have been widely discussed in academia, and great concerns over academic integrity have been reported worldwide~\cite{mulenga2024academic}. Students’ overreliance on such tools may significantly affect their critical thinking, problem-solving, communication (i.e., interaction with teachers and peers), and collaboration capabilities. Similarly, these tools have also resulted in growing concerns over the reliability and authenticity of content in various application domains, such as journalism \cite{jones2023generative}, art \cite{vskiljic2021art}, and law enforcement. For instance, such tools have significantly increased the creation and propagation of false information and fake news. These fake news and misinformation could affect the policies \cite{silva2024disinformation}, trust in governments, and cause chaos \cite{awwal2020examining}, which may compromise public safety. 

The encouraging point is that there are already some AI and Natural Language Processing (NLP)-based techniques that could automatically detect content (text) generated through such tools; however, such algorithms/tools can be easily bypassed by slight changes to auto-generated text (i.e., giving a human touch to the text). Interestingly, there are also AI-based tools for humanizing machine-generated text, making machine-generated text detection more challenging. Moreover, academia is also facing another important challenge in terms of authorship detection, i.e., verifying that the assignments, essays, project reports, and theses are written by the students themselves and are not produced by another human (e.g., peers or external experts). 
This challenge is equally important and harmful to academic integrity and learning outcomes. Considering the buzz generated by AI-based content-generating tools in academia and the consequences of unethical and fraudulent practices in the education sector, a quick response to these challenges from the research community is essential.  

The literature already reports some initial efforts to solve these issues. However, most of the existing solutions rely on training ML algorithms for differentiating between machine and human text, which could easily be bypassed by humanising machine-generated text. Moreover, the domain also lacks a large-scale benchmark dataset. To overcome these limitations, we propose a stylometry-based solution and produce a large-scale benchmark dataset. The styloetry-based solution helps in tackling multiple challenges faced by academia in ensuring academic integrity, for instance, it can also verify if a document is produced by the same author or if some other author/authors (human or machine) contributed to it. In this work, we tackle four tasks, including (i) differentiating between machine- and human-generated text, (ii) classification of single- and multi-authored documents, (iii) identifying the parts (paragraphs) of multi-authored documents where the author changes, and (iv) identification of the author for a portion of text in a multi-authored document. These four tasks are very critical in the educational context. For example, the differentiation between machine and human-generated text will avoid AI-based solutions to assessments. The second task predicts potential collaboration (illegal or legal) in the production of educational content. Finally, the last two tasks enable the identification of the parts produced by another author. Style analysis mainly involves the identification of distinct patterns in writing style by utilising intrinsic features in a single-authored or collaboratively authored document. Style change detection (SCD) is one of the key applications of style analysis that could be utilised for intrinsic plagiarism detection (IPD) \cite{manzoor2025stylometry}. IPD differs from external plagiarism detection, where a document is checked against existing sources or references. On the other hand, IPD aims to detect plagiarism by examining only the input document, determining whether parts of it are not from the same author, which could be a human or an AI-based tool. IPD leverages the writing style consistency of the authors throughout the document for plagiarism detection, independent of any other reference corpus, which makes it a more effective solution for tackling these challenges. Style change analysis is not new; however, it has been mainly explored in the context of human authors only. We believe such a detailed analysis will provide a baseline for future work in the domain. 

The key contributions of this work can be summarised as follows:

\begin{itemize}
    \item The paper proposes a novel style change analysis-based NLP framework for tackling some of the key challenges faced by academia in the era of generative AI. These tasks include differentiating between single- and multi-authored documents, classifying machine and human-generated text, and identifying the paragraphs where author changes occur in a multi-authored document.  
    \item The paper also analyzes the impact of cleverly crafted prompts on the performance of state-of-the-art NLP models-based tools for plagiarism and document authentication
    \item Two different sets of machine text are generated and embedded in a human-produced dataset, providing a valuable source for further research in the domain. The dataset and the source codes are made publicly available.
    \item The paper highlights key challenges associated with the topic and hints at some potential solutions.
\end{itemize}

The rest of the paper is organized as follows. Section \ref{sec:related_Work} provides an overview of the literature. Section \ref{sec:methodology} describes the proposed methodology. Section \ref{sec:experimental_results} discusses the experimental setup, conducted experiments, and the experimental results. Section \ref{sec:conclusion} concludes the paper.

\section{Related Work}
\label{sec:related_Work}
Considering the growing global concerns on document authentication and academic integrity in the era of generative AI, a keen interest has been observed in machine text generation and analysis recently \cite{iqbal2022survey,djeric2025exploring}. The literature reports several interesting studies exploring different aspects of the domain. One of the key aspects of machine text analysis is the classification of human and machine text, and a vast majority of the literature on the topic focuses on automatically differentiating between machine and human text \cite{fariello2025distinguishing}. This task is mostly treated as a binary classification problem, where various techniques and strategies are applied \cite{wu2025survey}. For instance, Bahad et al. \cite{bahad2024fine} fine-tuned a pre-trained model RoBERTa~\cite{liu2019roberta}, to classify AI- and human-generated text covering various topics. Ji et al. \cite{ji2024detecting} approached the task as a ternary classification problem by introducing a third category of undecided text samples that could belong to either class (i.e., human or AI). The third category was introduced to highlight the difficulties and limitations of the models in classifying text, emphasizing the need for clear and meaningful explanations to users. The AI-generated text was produced using four state-of-the-art models, including GPT-4.0~\cite{achiam2023gpt}, Google’s Gemini Pro~\cite{team2023gemini}, LLaMA 3.370B~\cite{dubey2024llama}, and Qwen2\-72B~\cite{bai2025qwen2}.  

Another important but less explored aspect of this research is the so-called stylometry analysis \cite{michailidis2022scientometric}, which is one of the key applications of text processing. Zamir et al. \cite{zamir2023document}, in their work on document authentication and provenance, hinted at the use of stylometry analysis, particularly Style Change Detection (SCD) for this task. Along with differentiating between machine and human text, SCD analysis could also address other relevant challenges faced by academia in terms of authorship detection, i.e., verifying that the assignments, essays, project reports, and theses are written by the students themselves and are not produced by another human (peers or other experts). The literature already reports several interesting studies on SCD and analysis, exploring different aspects and applications of the concept. One of the key applications of style change analysis is intrinsic plagiarism detection (IPD) \cite{manzoor2025stylometry}. In contrast to external plagiarism, where a document is checked against existing sources/references, IPD aims at discovering plagiarism by examining only the input document. Subsequently, decides whether parts of the input document are not from the same author, which could be a human or an AI-based tool. IPD leverages the writing style consistency of the authors over the length of the document for plagiarism detection, independent of any other reference corpus \cite{oberreuter2013text}. SCD could also be used for document authentication \cite{zamir2023document}. For instance, Zamir et al. \cite{zamir2024stylometry} used SCD for the identification of multi-authored documents as well as the text segments where author changes occurred 
Several other works also used stylometry analysis for the classification of single and collaboratively generated documents \cite{zangerle2021pan21,strom2021multi}. SCD has also been part of a benchmark competition, PAN \cite{kestemont2018overview,zangerle2020overview}, and has gained a lot of interest from the community.   
The literature already reports some interesting works on stylometry analysis; however, several aspects of stylometry in general and SCD analysis in particular have yet to be explored. The majority of the existing literature is based on human-authored documents only. There are very few recent works that use style change analysis to differentiate between human- and machine-authored documents on general topics. For instance, Kumar et al. \cite{kumarage2023stylometric} employed stylometry analysis to detect machine-generated text in Twitter posts, relying on traditional stylometry features such as phraseology, punctuation, and linguistic diversity. Opara et al. \cite{opara2024styloai} also relied on traditional stylometry features for the classification of human- and machine-generated text. More recently, several works have explored the feasibility of transformer-based models for stylometry analysis, providing contextual embeddings for words, sentences, or entire documents. These models have been proven to be very effective in different applications of stylometry analysis, such as author attribution \cite{fabien2020bertaa}, authorship detection \cite{almutairi2023bibert}, and SCD \cite{zamir2024stylometry}. 

The existing literature has mostly explored the first task (i.e., classification of human- and machine-generated text) using machine-generated text only. The other three tasks have been explored only in the context of human authors. Moreover, the literature lacks a benchmark dataset containing human- and machine-generated text, which would enable a detailed analysis of SCD by providing annotations for various relevant tasks. Therefore, in this work, we analyze four different tasks of SCD analysis by generating corresponding machine-text datasets under two different prompts.

\section{Methodology}
\label{sec:methodology}
\begin{figure}[]
\centering
\includegraphics[width=0.70\textwidth]{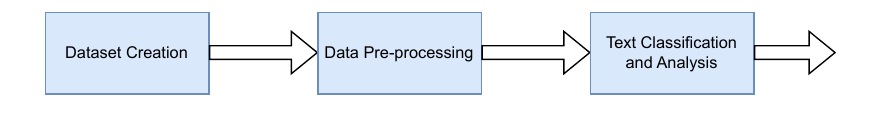}
\caption{The proposed methodology.}
	\label{fig:methodology}
\end{figure}

Figure \ref{fig:methodology} provides a block diagram of the proposed methodology, which can be mainly divided into three components. These include (i) dataset creation, (ii) data pre-processing, and (iii) text classification and analysis. At the start, we developed our dataset by generating machine-generated text through intelligent prompting. After developing the dataset, several pre-processing techniques were applied to handle various issues with the data, preparing it for the text classification and analysis using state-of-the-art NLP algorithms. In the final phase, we employ various NLP algorithms for analysing and differentiating between machine- and human-generated text in single- and
multi-authored documents. In the next subsections, we provide a detailed overview of each of the steps.

\subsection{Dataset Creation}
\begin{figure}[]
\centering
\includegraphics[width=0.6\textwidth]{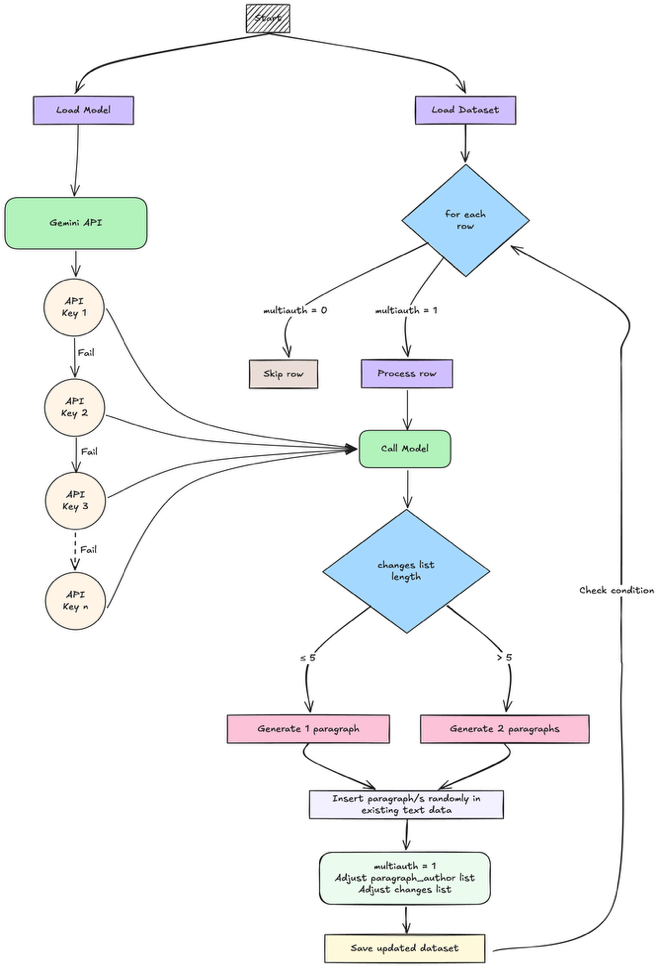}
\caption{Flowchart of the Data generation process. The same flowchart is used for both datasets, with only differences in the instruction sets.}
	\label{fig:data}
\end{figure}

In this phase of the methodology, we generated a large collection of machine-generated text paragraphs using the Gemini model (gemini-1.5-flash-latest) under two different prompts. The newly AI-generated text paragraphs were then inserted into the existing human-generated dataset at random locations for training and evaluation of our models for the four tasks. Figure \ref{fig:data} presents the flowchart of the dataset generation process. As shown in the figure, the text was generated using multiple free API keys.

The data generation process was designed according to the structure of the existing dataset, which contains text from human authors \cite{zangerle2021pan21}. In the human text dataset, text samples are organized into single and multi-authored files. Moreover, the dataset contains text samples of four authors. In our case, we consider the multi-authored files and generate machine text by considering AI/LLMs as the fifth author. Thus, as shown in the figure, for each sample, we first determine whether it is a single-authored or multi-authored text and proceed with the rest of the process if the sample is a multi-authored file. This approach helped avoid repetition and the labour involved in the text generation process. We note that skipping the single-authored files does not affect the machine vs human text classification task, as the machine-generated text, which is considered as the fifth author, is combined for that experiment.

In the multi-authored documents, the number of authors as well as the number of paragraphs varied, some containing fewer and others more. Keeping this in view and maintaining a reasonable ratio between the machine-generated and human-generated text, we add either a single paragraph or multiple paragraphs, depending on the number of paragraphs in each document. Thus, for each multi-authored document, we check if the number of paragraphs (changes list-end of a paragraph in a document) is less than or equal to, or greater than 5, we insert a single paragraph and a couple of machine-generated paragraphs, respectively. After successfully generating and inserting the machine text randomly, the author's change list is updated accordingly, which serves as a ground truth in some of the tasks. 

For the machine text, we generated two different prompts:  a normal one and a stricter one. In the normal prompt, we targeted a broader range of topics covered in the human-text benchmark dataset. The topics were extracted through modeling techniques, resulting in a list of topics extracted from the human text. For topic modeling, we used three different topic modeling algorithms, namely (i) BERTopic \cite{grootendorst2022bertopic}, Latent Dirichlet allocation (LDA) \cite{chauhan2021topic}, and Non-negative Matrix Factorization (NMF) \cite{lee2000algorithms}. These algorithms produced a list of topics extracted from the human text. We selected the top ten topics extracted by each algorithm, which were then fed into the Gemini model to generate extended topics by adding domains and subdomains of the topics. The extended topics were then fed into the Gemini model as part of the prompt for machine-generated text. Table \ref{tab:topics} provides the list of topics used in the prompt for text generation. In the \textbf{strict} prompt, we did not use pre-defined topics; rather, the model provided the human sample and asked to generate text in the same context. A comparison of the instructions used in both prompts is provided in \ref{tab:prompt_comparison}.


\begin{table}[]
\begin{tabular}{|l|l|}
\hline
 File Systems \& Data Management& Programming \& Software Development \\ \hline
Networking \& Cybersecurity & Cloud \& Virtualization Technologies \\ \hline
AI, ML \& Data Science & Hardware \& System Performance \\ \hline
Software Installation \& Troubleshooting & Enterprise IT \& DevOps \\ \hline
Tech Industry Trends & User Experience \& HCI \\ \hline
\end{tabular}
\caption{A list of the topics used in the prompt for text generation.}
	\label{tab:topics}
\end{table}
\begin{table}[]
\begin{tabular}{|p{7.4cm}|p{7.4cm}|}
\hline
\textbf{Instructions used in the Normal Prompt} & \textbf{Instructions used in the Strict Prompt} \\ \hline
\begin{itemize}
\item Generate a single paragraph, focusing on any of the following topics within the domains: File Systems \& Data Management, Programming \& Software Development, Networking \& Cybersecurity, Cloud \& Virtualization Technologies, AI, ML \& Data Science, Hardware \& System Performance, Software Installation \& Troubleshooting, Enterprise IT \& DevOps, Tech Industry Trends, User Experience \& HC
\item  Do not include any introductory text, explanations, prefaces, or meta information such as ’Here is a new paragraph.’.
  \item Ensure variety across the five generated paragraphs.
    \item Each paragraph should be between 40 and 80 words.
    \item Cover features, principles, challenges, adoption trends, and pros/cons.
    \item Use an informative and professional tone with factual correctness.
    \item Avoid redundancy or vague speculation.
\end{itemize}
& \begin{itemize}
   \item Based on the context of the text in the single quotes, generate a new paragraph in which several words will equal the average words in the other provided paragraphs
    \item Only output the paragraph; do not include any introductory text, explanations, prefaces, or meta information such as 'Here is a new paragraph.'.
   \item  Avoid extra labels, annotations, or machine-generated markers in the output.
     Do not summarize or directly use ideas from the provided paragraphs.
\item Ensure the paragraph is relevant to the topic and adds a new perspective or insight.
\item Do not include introductory or explanatory text—produce only the paragraphs directly.
\item Focus on contributing meaningful content that complements the general agenda or subject 
matter.
\item Pay attention to the length of your generated paragraph, ensuring it matches the average word count of the other paragraphs.
\end{itemize}
\\
\hline
\end{tabular}
\caption{A comparison of the set of instructions in the prompt for the generation of machine text with the model.}
	\label{tab:prompt_comparison}
\end{table}


\subsection{Data Pre-processing}
The data was analyzed and cleaned before being fed into the models. Various tasks were carried out during data pre-processing, including data cleaning, duplicate detection and removal, and data balancing. Firstly, the dataset was checked for null values and duplicate entries by analyzing the text and authors' information. This was followed by data balancing using class weights. We note that during this phase, we explored various data balancing techniques, including oversampling, undersampling, SMOTE~\cite{chawla2002smote}, SMOTEENN~\cite{batista2004study}, ADASYN~\cite{he2008adasyn}, NearMiss, Tomek, and NearMiss-Tomek. However, the final approach was based on class weights by assigning higher weights to the minority class to pay more attention to them. These weights were incorporated into the cross-entropy loss function so that 
minority-class errors carry a greater penalty.

\subsection{Text Classification}
For text classification, we utilised four state-of-the-art models: BERT (base model), ALBERT, DistilBERT, and RoBERTa. Our choice of these models is based on their proven performance on similar tasks \cite{konuma2025japanese}. The lower/intermediate layers of these models extract more traditional stylometry features, such as function words, punctuation, and sentence rhythm, etc. The BERT-base model consists of 12 transformer layers and attention heads, with a token length of 512. DistilBERT is a variant of the base model, providing a smaller and faster model without significantly impacting performance. RoBERTa is another variant of the base model with a different training strategy, including a larger training set and time, as well as the removal of next-sentence prediction. All the models are fine-tuned on the new dataset using different dropout and attention dropout rates to achieve better results. More details of the hyperparameters are provided in Section \ref{sec:experimental_results}.

\section{Experiments and Results}
\label{sec:experimental_results}

\subsection{Experimental Setup}
To achieve the multi-faceted objectives of this work, we conducted multiple experiments, exploring different aspects of the application. These experiments include:

\begin{itemize}
     \item \textbf{Classification of machine and human-generated text}: This is a binary classification task where the models predict whether a document was produced by a human author or a machine. 
    \item \textbf{Classification of single and multi-authored documents}: This is also a binary classification task where the models predict whether a document was produced by a single author or is multi-authored. The single-authored documents include text produced by humans and machines. In the case of multi-authored documents, machine-generated text was randomly embedded in the documents produced by multiple human authors.
    \item \textbf{Author change detection}: This is also a binary classification task that involves the identification of subsequent paragraphs where the author changes. Each pair of subsequent paragraphs in multi-authored documents is labeled either 0 or 1, representing ''no changes'' and ''author changes'', respectively.
    \item \textbf{Author recognition}: In this experiment, we need to identify the author for each document that could be either a single-authored or collaboratively written document. Thus, this task is treated as a multi-label classification task with five different authors, including the machine/AI as one of the authors.
\end{itemize}

During the experiments, we used various techniques and strategies. For instance, considering our limited computational resources, we used the Gradient accumulation technique, allowing us to use a larger batch size. The batch size used in the experiments was 32; however, the Gradient accumulation of 2 steps effectively made it 64. The learning rate was initialized at $0.00001$, decaying via a cosine scheduler with 200 warmup steps.  Similarly, the weight decay was used during training, preventing overfitting and improving the generalization of models. Moreover, we run the validation every 500 steps and save the best 
model checkpoint based on the weighted F1-score. We also enabled the mixed precision (fp16) training to speed up computation and reduce memory usage. Logging was employed, recording metrics every 50 steps and retaining only the five most recent checkpoints, to monitor the progress. Moreover, we used 70\%, 15\%, and 15\% samples for training, validation, and testing.

\begin{table}[]
\begin{tabular}{|c|c|c|c|}
\hline
\textbf{Hyper-parameter} & \textbf{Value} & \textbf{Hyper-parameter} & \textbf{Value} \\ \hline
Batch Size & 32 & Gradient Accumulation & 2  \\ \hline
Learning Rate & 0.00001  & Weight Decay &  0.01 \\ \hline
 Epochs&  5 & Logging & 50 \\ \hline
\end{tabular}
\caption{A summary of the hyperparameters used during the experiments.}
	\label{tab:parameters}
\end{table}

\subsection{Experimental Results}

\subsubsection{Machine and Human-text Classification}
Table \ref{tab:human_machine} presents the experimental results of the first task with text generated using both the normal and the strict prompt in terms of accuracy, precision, recall, and F1-score. Overall, the results are very good on both datasets, demonstrating the capabilities of the transformer-based NLP models in differentiating between human- and machine-generated text. As expected, the results are significantly better on the dataset, including AI-generated text through the normal prompt, compared to the strict one. This demonstrates the complexity of differentiating human text from AI text when generated through cleverly crafted prompts. Regarding the performance of the individual models, no significant differences are observed in the results (all metrics) of the models on both datasets. Moreover, the precision and recall are in the same range, demonstrating a balanced trade-off between correctly predicting the positive and negative samples. We also analysed the performance of these models on individual classes (i.e., the classification of AI and human text). We observed a higher difference in the models' performances on the individual classes in the case of the strict prompt compared to the normal one, where scores were higher for the human text compared to the AI-generated text.
\begin{table}[]
\begin{tabular}{|c|cccc|cccc|}
\hline
\multirow{2}{*}{\textbf{Model}} & \multicolumn{4}{c|}{\textbf{Normal Prompt}} & \multicolumn{4}{c|}{\textbf{Strict Prompt}} \\ \cline{2-9} 
 & \multicolumn{1}{c|}{Accuracy} & \multicolumn{1}{c|}{Precsion} & \multicolumn{1}{c|}{Recall} & F1-score & \multicolumn{1}{c|}{Accuracy} & \multicolumn{1}{c|}{Precsion} & \multicolumn{1}{c|}{Recall} & F1-score \\ \hline
distilbert-base-uncased & \multicolumn{1}{c|}{0.9998} & \multicolumn{1}{c|}{0.9998} & \multicolumn{1}{c|}{0.9998} & 0.9998 & \multicolumn{1}{c|}{0.9477} & \multicolumn{1}{c|}{0.9463} & \multicolumn{1}{c|}{0.9477} & 0.9463 \\ \hline
Albert-base-v2 & \multicolumn{1}{c|}{0.9995} & \multicolumn{1}{c|}{0.9995} & \multicolumn{1}{c|}{0.9995} & 0.9995 & \multicolumn{1}{c|}{0.9456} & \multicolumn{1}{c|}{0.9440} & \multicolumn{1}{c|}{0.9456} & 0.9441 \\ \hline
 Roberta-base & \multicolumn{1}{c|}{0.9997} & \multicolumn{1}{c|}{0.9997} & \multicolumn{1}{c|}{0.9997} & 0.9997 & \multicolumn{1}{c|}{0.9487} & \multicolumn{1}{c|}{0.9473} & \multicolumn{1}{c|}{0.9487} & 0.9474 \\ \hline
 Bert-base-uncased  & \multicolumn{1}{c|}{0.9998} & \multicolumn{1}{c|}{0.9998} & \multicolumn{1}{c|}{0.9998} & 0.9998 & \multicolumn{1}{c|}{0.9483} & \multicolumn{1}{c|}{0.9469} & \multicolumn{1}{c|}{0.9483} & 0.9467 \\ \hline
\end{tabular}
\caption{Experimental results on the classification of machine and human text on both datasets.}
	\label{tab:human_machine}
\end{table}

\begin{figure}[]
\centering
\includegraphics[width=0.8\textwidth]{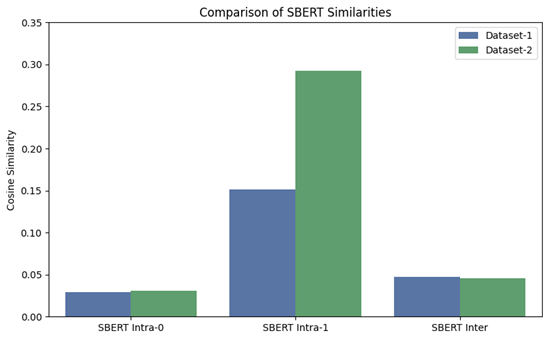}
\caption{A comparison of SBERT similarities of both datasets. Dataset 1 represents the dataset generated with the normal prompt, while Dataset 2 contains machine-generated text using the strict prompts. Moreover, class 0 and class 1 represent human and machine text, respectively.}
	\label{clusters}
\end{figure}

To further investigate the differences in model performance on both datasets, a comparative semantic analysis is conducted in terms of inter-class separation and intra-class cohesion using Sentence-BERT Embeddings. Such analysis, using inter-class separation and intra-class cohesion, is crucial for any classification task, particularly in the context of a machine-generated dataset, as it highlights how well each dataset preserves semantic organization. The average cosine similarity between class 0 (human text) and class 1 (machine text) of the dataset generated with the normal prompt embeddings is very low ($0.047$). In the embedding space, these two classes formed completely separate clusters. A similar trend is observed in the case of the dataset generated with the strict prompt, with the same inter-class similarity shown in Figure \ref{clusters}. However, a different trend is observed in the case of intra-class cohesion, which is another critical factor affecting the classification performance on a dataset. In terms of intra-class cohesion, we noticed that in the dataset generated with the strict prompt, class 1 examples were moderately similar to each other, having an average intra-class cohesion of $0.14$, while class 0 samples were found to be quite diverse with an average intra-class cohesion value of $0.03$. We note that class 0 is human-generated text, and it remained the same for both datasets. Thus, we computed the intra-class cohesion for the machine-generated class only on dataset 2, which is generated using the normal prompt. The average intra-class cohesion for class 1 of dataset 2 is comparatively high ($0.29$), demonstrating more compactness/similarity in the machine-generated samples. The low intra-class cohesion of class 1 of dataset 1 is mainly due to the instructions used in the prompt, asking for strictly similar text to be produced in the same context of the human produced text in each document. 

The high intra-class cohesion of the machine-generated data resulted in better classification score on machine-generated text compared to the one generated with the strict prompt. This is also proved by the classification scores on the individual classes on both datasets. As a sample, we provide classification results of the Albert model in Table \ref{individual_Scores}. 

\begin{table}[]
\begin{tabular}{|c|ccc|ccc|}
\hline
\multirow{2}{*}{\textbf{Class/Author}} & \multicolumn{3}{c|}{\textbf{Dataset 1 (Strict Prompt)}} & \multicolumn{3}{c|}{\textbf{Dataset 2 (Normal Prompt)}} \\ \cline{2-7} 
 & \multicolumn{1}{c|}{Precsion} & \multicolumn{1}{c|}{Recall} & F1-score & \multicolumn{1}{c|}{Precsion} & \multicolumn{1}{c|}{Recall} & F1-score \\ \hline
Class 0 (Human Text) & \multicolumn{1}{c|}{0.9567} & \multicolumn{1}{c|}{0.9795} & 0.9780 & \multicolumn{1}{c|}{0.9995} & \multicolumn{1}{c|}{1.00} & 0.9998 \\ \hline
Class 1 (Machine Text) & \multicolumn{1}{c|}{0.8779} & \multicolumn{1}{c|}{0.7691} & 0.8199 & \multicolumn{1}{c|}{1.00} & \multicolumn{1}{c|}{0.9971} & 0.9985 \\ \hline
\end{tabular}
\caption{Classification scores of one of the models (Albert) on the individual classes of both datasets to demonstrate the impact of high intra-class cohesion of class 1 of dataset 2.}
	\label{individual_Scores}
\end{table}

\subsubsection{Classification of Single and Multi-authored Documents}
Table \ref{tab:single-multi} presents the experimental results on the classification of single and multi-authored documents. As can be seen, the performances of the models are very high compared to the previous task on the dataset generated with the strict prompt. The higher scores of the models reflect the lower complexity of the task, where the models were able to distinguish between single and multi-authored documents more accurately. As noted earlier, in this experiment, the single documents include both human- and machine-generated text, while the multi-authored documents comprise text from multiple human authors, as well as machine-generated text embedded in the documents at random locations. This performance improvement may be attributed to a comparatively lower involvement of AI-generated text compared to the first task of differentiating between human and machine text. 

\begin{table}[]
\begin{tabular}{|c|cccc|cccc|}
\hline
\multirow{2}{*}{\textbf{Model}} & \multicolumn{4}{c|}{\textbf{Normal Prompt}} & \multicolumn{4}{c|}{\textbf{Strict Prompt}} \\ \cline{2-9} 
 & \multicolumn{1}{c|}{Accuracy} & \multicolumn{1}{c|}{Precsion} & \multicolumn{1}{c|}{Recall} & F1-score & \multicolumn{1}{c|}{Accuracy} & \multicolumn{1}{c|}{Precsion} & \multicolumn{1}{c|}{Recall} & F1-score \\ \hline
 Distilbert-base-uncased& \multicolumn{1}{c|}{0.9866} & \multicolumn{1}{c|}{0.9869} & \multicolumn{1}{c|}{0.9866} & 0.9866 & \multicolumn{1}{c|}{0.9845} & \multicolumn{1}{c|}{0.9844} & \multicolumn{1}{c|}{0.9845} & 0.9844 \\ \hline
Albert-base-v2 & \multicolumn{1}{c|}{0.9841} & \multicolumn{1}{c|}{0.9850} & \multicolumn{1}{c|}{0.9841} &0.9842  & \multicolumn{1}{c|}{0.9853} & \multicolumn{1}{c|}{0.9854} & \multicolumn{1}{c|}{0.9853} & 0.9853 \\ \hline
Roberta-base  & \multicolumn{1}{c|}{0.9924} & \multicolumn{1}{c|}{0.9925} & \multicolumn{1}{c|}{0.9924} & 0.9925 & \multicolumn{1}{c|}{0.9824} & \multicolumn{1}{c|}{0.9825} & \multicolumn{1}{c|}{0.9824} & 0.9822 \\ \hline
 Bert-base-uncased  & \multicolumn{1}{c|}{0.9857} & \multicolumn{1}{c|}{0.9862} & \multicolumn{1}{c|}{0.9857} & 0.9858 & \multicolumn{1}{c|}{0.9774} & \multicolumn{1}{c|}{0.9773} & \multicolumn{1}{c|}{0.9774} & 0.9772 \\ \hline
\end{tabular}
\caption{Experimental results on the single and multi-authored documents classification task using the normal and strict prompts for AI-text generation.}
	\label{tab:single-multi}
\end{table}

\subsubsection{Author Change Detection in Multi-authored Documents}
Table \ref{tab:change_detection} presents the experimental results of the author change detection task. This task involves predicting the author changes in subsequent paragraphs (i.e., whether the next paragraph is produced by the same author or not). The task involves five authors, including four human authors. The performances of the models dropped significantly compared to the previous tasks. The highest F1-scores obtained on the datasets generated with normal and strict prompts are 0.680 and 0.70, respectively. The lower performance is mainly due to the complexity of the task, as also confirmed by the literature \cite{zamir2024stylometry}. To further explore the experimental results, we analyzed the per-class scores and noticed better performances on the positive class, where the author changes were correctly predicted most of the time. However, some misclassification cases were also observed on class 0, representing the samples where the author did not change.

\begin{table}[]
\begin{tabular}{|c|cccc|cccc|}
\hline
\multirow{2}{*}{\textbf{Model}} & \multicolumn{4}{c|}{\textbf{Normal Prompt}} & \multicolumn{4}{c|}{\textbf{Strict Prompt}} \\ \cline{2-9} 
 & \multicolumn{1}{c|}{Accuracy} & \multicolumn{1}{c|}{Precsion} & \multicolumn{1}{c|}{Recall} & F1-score & \multicolumn{1}{c|}{Accuracy} & \multicolumn{1}{c|}{Precsion} & \multicolumn{1}{c|}{Recall} & F1-score \\ \hline
Albert-base-v2 & \multicolumn{1}{c|}{0.674} & \multicolumn{1}{c|}{0.6982} & \multicolumn{1}{c|}{0.674} &0.6806  & \multicolumn{1}{c|}{0.6818} & \multicolumn{1}{c|}{0.6964} & \multicolumn{1}{c|}{0.6818} & 0.6867 \\ \hline
 Bert-base-uncased& \multicolumn{1}{c|}{0.5878} & \multicolumn{1}{c|}{0.6795} & \multicolumn{1}{c|}{0.5878} & 0.5928 & \multicolumn{1}{c|}{0.64} & \multicolumn{1}{c|}{0.6908} & \multicolumn{1}{c|}{0.64} & 0.6484 \\ \hline
Distilbert-base-uncased  & \multicolumn{1}{c|}{0.6699} & \multicolumn{1}{c|}{0.7023} & \multicolumn{1}{c|}{0.6699} & 0.6774 & \multicolumn{1}{c|}{0.6973} & \multicolumn{1}{c|}{0.7147} & \multicolumn{1}{c|}{0.6973} & 0.7026 \\ \hline
Roberta-base   & \multicolumn{1}{c|}{0.6667} & \multicolumn{1}{c|}{0.6913} & \multicolumn{1}{c|}{0.6667} &0.6735  & \multicolumn{1}{c|}{0.6948} & \multicolumn{1}{c|}{0.7149} & \multicolumn{1}{c|}{0.6948} & 0.7006 \\ \hline
\end{tabular}
\caption{Experimental results on the author change detection task in multi-authored documents using AI-text generated through the normal and strict prompts.}
	\label{tab:change_detection}
\end{table}



\subsubsection{Author Recognition in Multi-authored Documents}
Table \ref{tab:author_recognition} presents the experimental results of the final experiment, where we predict the author/s of a document. It is a multi-label classification problem where multiple authors could have contributed to a document. The F1-scores on both datasets are very low compared to the other tasks. For the calculation of the F1-score for this task, we used optimal thresholds for each class, as these classes have different numbers of samples and different distributions. To find the optimal thresholds, we used a grid search approach by looping through and calculating the F1 score at different thresholds ranging from 0.05 to 0.95 with a 0.01 step size. The final thresholds used for classes 1 to 5 are $0.46, 0.48, 0.56, 0.51$, and $0.9$, respectively.

Another interesting observation is the significant differences in the precision and recall values produced by each model. The precision of each model is very low compared to the recall, which indicates that the models are over-predicting the positive samples. During our analysis, we noticed that the performance of the models is very low on certain classes. As a reference, per-class results are provided for the ALBERT-base model in Table \ref{tab:author_recognition_perclass}. The scores, especially the precision for the recognition of authors 2, 3, and 4, are very poor. One of the potential reasons for the lower performance on these classes is the low number of samples in these classes.


\begin{table}[]
\begin{tabular}{|c|ccc|ccc|}
\hline
\multirow{2}{*}{\textbf{Model}} & \multicolumn{3}{c|}{\textbf{Normal Prompt}} & \multicolumn{3}{c|}{\textbf{Strict Prompt}} \\ \cline{2-7} 
 & \multicolumn{1}{c|}{Precsion} & \multicolumn{1}{c|}{Recall} & F1-score & \multicolumn{1}{c|}{Precsion} & \multicolumn{1}{c|}{Recall} & F1-score \\ \hline
Albert-base-v2 & \multicolumn{1}{c|}{0.3756} & \multicolumn{1}{c|}{0.9106} & 0.5318 & \multicolumn{1}{c|}{0.3869} & \multicolumn{1}{c|}{0.7433} & 0.5089 \\ \hline
 bert-base-uncased& \multicolumn{1}{c|}{0.3564} & \multicolumn{1}{c|}{0.9314} & 0.5155 & \multicolumn{1}{c|}{0.3835 } & \multicolumn{1}{c|}{0.7306} & 0.503 \\ \hline
 distilbert-base-uncased & \multicolumn{1}{c|}{0.355} & \multicolumn{1}{c|}{0.9357} & 0.5147 & \multicolumn{1}{c|}{0.383} & \multicolumn{1}{c|}{0.7094} & 0.4974 \\ \hline
   Roberta& \multicolumn{1}{c|}{0.3561} & \multicolumn{1}{c|}{0.9369} & 0.516 & \multicolumn{1}{c|}{0.3842} & \multicolumn{1}{c|}{0.7344} & 0.5045 \\ \hline
\end{tabular}
\caption{Experimental results on the author recognition task in multi-authored documents, including AI-text generated through the normal and strict prompts.}
	\label{tab:author_recognition}
\end{table}






\begin{table}[]
\begin{tabular}{|c|ccc|ccc|}
\hline
\multirow{2}{*}{\textbf{Class/Author}} & \multicolumn{3}{c|}{\textbf{Normal Prompt}} & \multicolumn{3}{c|}{\textbf{Strict Prompt}} \\ \cline{2-7} 
 & \multicolumn{1}{c|}{Precsion} & \multicolumn{1}{c|}{Recall} & F1-score & \multicolumn{1}{c|}{Precsion} & \multicolumn{1}{c|}{Recall} & F1-score \\ \hline
Author 1 & \multicolumn{1}{c|}{0.5396} & \multicolumn{1}{c|}{0.999} & 0.70 & \multicolumn{1}{c|}{0.6483} & \multicolumn{1}{c|}{0.7676} & 0.7029 \\ \hline
Author 2 & \multicolumn{1}{c|}{0.2674} & \multicolumn{1}{c|}{0.9991} &  0.4212& \multicolumn{1}{c|}{0.3005} & \multicolumn{1}{c|}{0.7014} & 0.4208 \\ \hline
Author 3 & \multicolumn{1}{c|}{0.1542} & \multicolumn{1}{c|}{0.7297} &  0.2546 & \multicolumn{1}{c|}{0.2027} & \multicolumn{1}{c|}{0.5624} & 0.2980 \\ \hline
Author 4 & \multicolumn{1}{c|}{0.0964} & \multicolumn{1}{c|}{0.0086} & 0.0158 & \multicolumn{1}{c|}{0.0890} & \multicolumn{1}{c|}{0.5404} & 0.1528 \\ \hline
Author 5 & \multicolumn{1}{c|}{0.9990} & \multicolumn{1}{c|}{0.9960} & 0.9978 & \multicolumn{1}{c|}{0.8516} & \multicolumn{1}{c|}{0.7983} & 0.8241  \\ \hline
\end{tabular}
\caption{Experimental results of the Albert model on each class of both datasets for the author recognition task.}
	\label{tab:author_recognition_perclass}
\end{table}

\section{Conclusion and Future Work}
\label{sec:conclusion}
In this paper, we analyzed and proposed solutions for some of the key challenges associated with academic integrity in the era of generative AI in the form of plagiarism and authorship verification. The paper explored four challenges, including differentiating between machine-generated and human-generated text, identifying if the document is produced by a single or multiple authors, identification of paragraphs where the author switches, and the identification of authors for each paragraph of a multi-authored document. Two different sets of machine-generated text, embedded in a human-produced dataset, were used for the experiments, demonstrating the impact of the prompt (set of instructions used for generating the machine) on the AI-based plagiarism and content authentication tools. During the experiments, we observed a significant reduction in the performance of the models in predicting machine-generated text, demonstrating how such tools could be bypassed with cleverly crafted prompts. The better performance of the AI-based solutions for the classification of single and multi-authored documents is very encouraging in ensuring the authenticity of educational content, for instance, an assignment is produced by a single author. Similarly, identification of author changes and author recognition in a collaboratively produced document can also ensure academic integrity. However, the lower experimental results on these two tasks demonstrate the complexity of the tasks. 

Despite some initial efforts, several aspects of the domain are yet to be explored. For instance, the domain lacks a large-scale benchmark dataset in the educational context, exploring different aspects of various types of educational content, such as project reports, essays, theses, and other assignments. In this regard, multi-modal solutions, combining linguistic, semantic, contextual, and metadata/features, could be very effective, especially in the identification of author switching and recognition in collaboratively produced documents. Moreover, robust solutions, by exploring adversarial frameworks, would help in coping with challenges associated with the detection of AI educational content generated through cleverly crafted prompts. 

\section{Author contributions statement}
H. A, K. A., and M. A.A. conceived the experiments, M.A.A and Y.A. developed the software, M. A. A., and H. A. conducted the experiments, N.A., M. M. A., and N.A. analyzed the results. H.A, Y.A, M. M. A., and K.A. prepared the initial draft of the paper. A. M. A., A. A. O., N. A, M. M. A., and K. A. reviewed the manuscript. 

\section{Data availability}
The code and associated data are publicly available on GitHub~\url{https://github.com/MASIFAYUB/Stylometry-Analysis-of-Human-and-Machine#}, which can be used for research purposes. 

\section{Conflict of Interest}
The authors declare no conflict of interest.
\bibliographystyle{spmpsci}      
\bibliography{sigproc}

\end{document}